%% file: iros2019_main.tex
\documentclass[letterpaper, 10 pt, conference]{ieeeconf}  

\IEEEoverridecommandlockouts                              

\usepackage{graphics} 
\usepackage{epsfig} 
\usepackage{mathptmx} 
\usepackage{times} 
\usepackage{amsfonts} 
\usepackage{amssymb}  
\usepackage{amsmath} 

\usepackage{graphicx}
\usepackage{helvet}

\usepackage{url}
\usepackage{rotating}
\usepackage{xspace}
\usepackage{booktabs}       
\usepackage{nccmath}
\usepackage{nicefrac}       
\usepackage{microtype}      
\usepackage{comment}		
\usepackage{adjustbox}		
\usepackage{tabu}
\usepackage{calc}			
\usepackage{wrapfig}

\usepackage{multirow}
\usepackage{float}
\usepackage[hang,flushmargin]{footmisc}
\usepackage[neverdecrease]{paralist}
\usepackage{algorithm}
\usepackage{algorithmic}
\usepackage{calc}
\usepackage{color}
\usepackage{colortbl}

\newcommand{\Figref}[1]{Figure~\ref{#1}}
\newcommand{\Tabref}[1]{Table~\ref{#1}}
\newcommand{\Eqnref}[1]{Equation~(\ref{#1})}

\usepackage[pagebackref=false,breaklinks=true,colorlinks,citecolor=blue,linkcolor=blue,bookmarks=false]{hyperref}

\usepackage[mathcal]{eucal}

\include{def2}

\definecolor{CuGray}{gray}{0.9}
\newcolumntype{g}{>{\columncolor{CuGray}}c}

\makeatletter
\DeclareRobustCommand\onedot{\futurelet\@let@token\@onedot}
\def\@onedot{\ifx\@let@token.\else.\null\fi\xspace}

\newcommand{\dashrule}[1][black]{%
  \color{#1}\rule[\dimexpr.5ex-.2pt]{4pt}{.4pt}\xleaders\hbox{\rule{4pt}{0pt}\rule[\dimexpr.5ex-.2pt]{4pt}{.4pt}}\hfill\kern0pt%
}

\newcommand*{\Scale}[2][4]{\scalebox{#1}{$#2$}}%

\def\eg{\emph{e.g}\onedot} 
\def\ie{\emph{i.e}\onedot}

\def\etal{\emph{et al}\onedot}

\title{\LARGE \bf
Learning Residual Flow as Dynamic Motion from Stereo Videos
}

\author{Seokju Lee$^{\dagger,1}$, Sunghoon Im$^{1}$, Stephen Lin$^{2}$ and In So Kweon$^{1}$
\thanks{$^\dagger$Part of this work was done during an internship at Microsoft Research Asia}
\thanks{$^{1}$S. Lee, S. Im and I. S. Kweon are with the Robotics and Computer Vision Laboratory, KAIST, Daejeon, 34141, Republic of Korea
        {\tt\small{\{snapillar, dlarl8927, iskweon77\}}@kaist.ac.kr}}%
\thanks{$^{2}$S. Lin is with the Microsoft Research Asia, Beijing, 100080, China
        {\tt\small stevelin@microsoft.com}}%
}

\begin{document}

\maketitle
\thispagestyle{empty}
\pagestyle{empty}

\begin{abstract}

We present a method for decomposing the 3D scene flow observed from a moving stereo rig into stationary scene elements and dynamic object motion. Our unsupervised learning framework jointly reasons about the camera motion, optical flow, and 3D motion of moving objects. Three cooperating networks predict stereo matching, camera motion, and residual flow, which represents the flow component due to object motion and not from camera motion. Based on rigid projective geometry, the estimated stereo depth is used to guide the camera motion estimation, and the depth and camera motion are used to guide the residual flow estimation. We also explicitly estimate the 3D scene flow of dynamic objects based on the residual flow and scene depth. Experiments on the KITTI dataset demonstrate the effectiveness of our approach and show that our method outperforms other state-of-the-art algorithms on the optical flow and visual odometry tasks.


\end{abstract}

\input{sec_1}

\input{sec_2}

\input{sec_3_v2}

\input{sec_4}

\section{CONCLUSION}

We have presented a joint learning framework that predicts camera motion, optical flow, and the 3D motion of moving objects in an unsupervised manner. Leveraging a stereo vision system, the proposed pose network and residual flow network are jointly trained with a new multi-phase optimization technique, which improves the quality of all the individual flow components (static, moving, and all) and 6-DoF camera poses. We have demonstrated that the proposed method based on a stereo vision system effectively decouples the motion sources, namely camera motion and object motion. Experiments on the KITTI benchmark dataset show that our method outperforms state-of-the-art methods on optical flow and visual odometry. In future, we will extend this work to the dynamic region segmentation and generalized 3D motion prediction of generic rigid body objects.

\textbf{Acknowledgement}~~
This work was supported by Institute for Information \& communications Technology Promotion (IITP) grant funded by the Korea government (MSIT, 2017-0-01780).

{\small
\bibliographystyle{IEEEtran}
\bibliography{egbib}
}

\end{document}

%% file: def2.tex
\newcommand{\calL}{{\mathcal{L}}}

\newcommand{\be}{\begin{eqnarray}}
\newcommand{\ee}{\end{eqnarray}}
\newcommand{\bee}{\begin{eqnarray*}}
\newcommand{\eee}{\end{eqnarray*}}

\newcommand{\matrixb}{\left[ \begin{array}}
\newcommand{\matrixe}{\end{array} \right]}

%% file: sec_1.tex
\section{INTRODUCTION}

Interpretation of both agent status and scene change is a fundamental problem in robotics. The ability to infer the motion of cameras and objects enables robots to reason and operate in dynamic real-world, such as autonomous navigation~\cite{geiger20143d,shashua2004pedestrian}, dynamic scene reconstruction~\cite{newcombe2015dynamicfusion}, and robot manipulation~\cite{byravan2017se3}. Recent developments in deep neural networks (DNNs) have led to the state-of-the-art performance in scene geometry reconstruction using monocular images~\cite{eigen2014depth,garg2016unsupervised} and stereo images~\cite{mayer2016large,chang2018pyramid}, optical flow~\cite{dosovitskiy2015flownet,sun2018pwc}, and scene flow estimation~\cite{lv2018learning}. In addition, there has been significant progress in learning-based camera localization~\cite{zhou2017unsupervised,zhou2018deeptam}. However, the joint estimation of scene geometry, scene flow and ego-motion is still an open and challenging problem, especially in complex urban environments that contain many dynamic objects. 

To tackle this problem, we design an unsupervised learning framework that includes depth, pose and residual flow networks as shown in~\Figref{fig_teaser}. In contrast to the previous monocular-based approaches~\cite{yin2018geonet,zou2018df,yang2018every,ranjan2019collaboration}, we introduce a stereo-based motion and residual flow learning module to handle non-rigid cases caused by dynamic objects in an unsupervised manner. Employing a stereo-based system enables to solve the ambiguity of 3D object motion in the monocular setting, and to obtain reliable depth estimates that can facilitate the other tasks. We leverage the pretrained stereo matching network to jointly train a pose and a residual flow network. We formulate the problem of rigid optical flow estimation $F^{rig}$ as a geometric transformation of the estimated depth and camera poses. Then, we estimate the moving object flows $F^{res}$ as a residual component to the rigid flow as
\begin{equation}
\Scale[0.99]
{
\begin{aligned}
F^{tot} = F^{rig} + F^{res},
\end{aligned}
}
\end{equation}
where $F^{tot}$ denotes the total optical flow between consecutive frames. Finally, we show that the proposed pipeline obtains the 3D motion of dynamic objects, as well as the scene geometry and agent motion. The distinctive points of our approach are summarized as follows:

\begin{figure}[t] 
	\centering
    \includegraphics[width=1\linewidth]{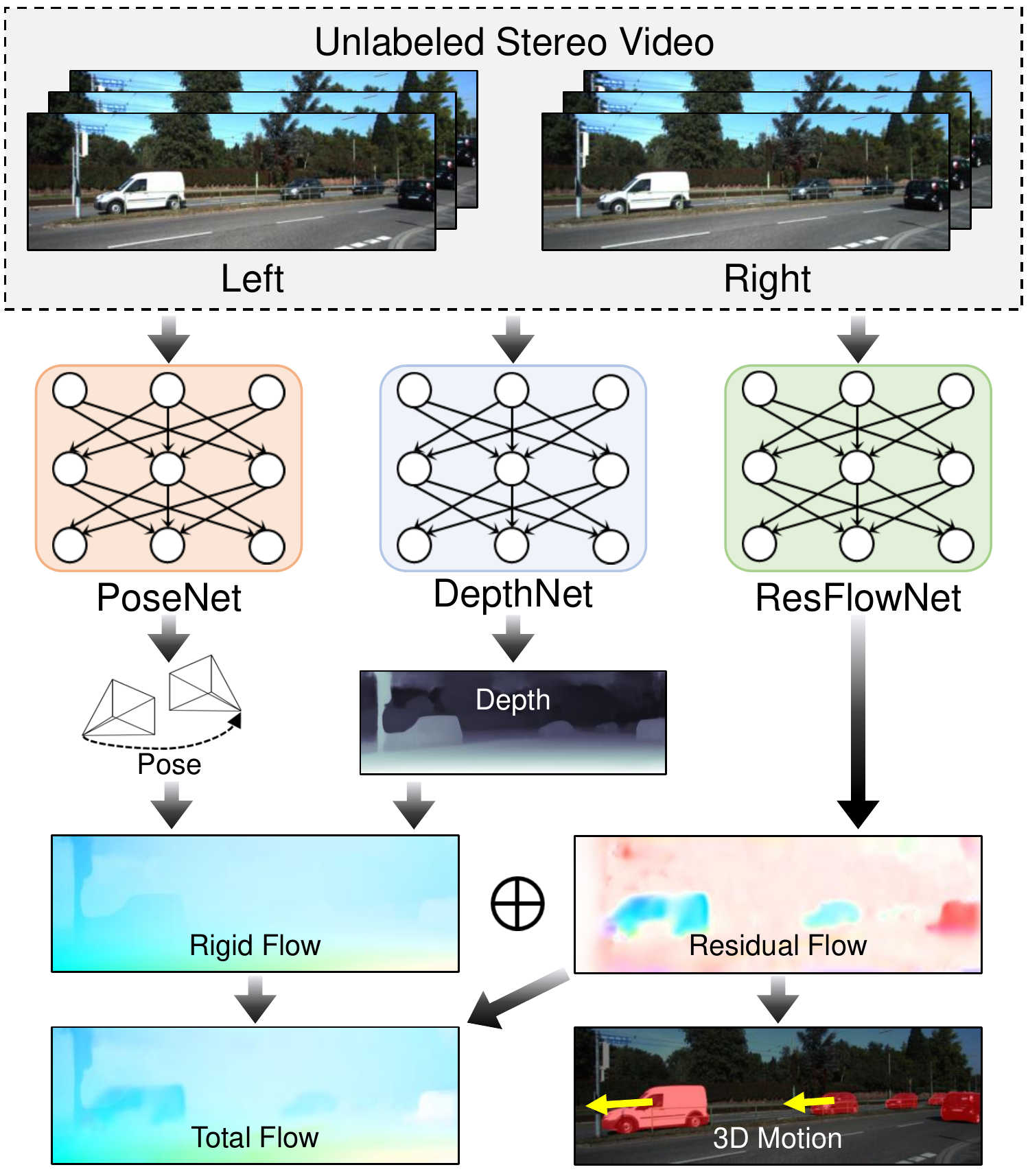}
	\caption{Our proposed system learns scene flow and visual odometry with unlabeled stereo videos. Based on the residuals of rigid and non-rigid flow, our system decouples the motions of static and dynamic objects effectively.}
	\label{fig_teaser}
\end{figure}

\begin{enumerate}
\item~We propose a novel end-to-end unsupervised deep neural network that simultaneously infers camera motion, optical flow, and 3D motions of moving objects using a stereo vision system. 
\item~We explicitly reason about 3D motion flows of dynamic objects based on the residuals of rigid and non-rigid motion flow. We demonstrate that the depth, pose, and flow estimation networks play complementary roles.
\item~We show that the proposed scheme achieves state-of-the-art results in optical flow and visual odometry on the KITTI benchmark dataset.
\end{enumerate}

%% file: sec_2.tex
\section{RELATED WORK}

\subsection{Depth from a single image} \quad
Recently, methods have emerged for inferring scene depth from a single image. Eigen~\etal~\cite{eigen2014depth} first demonstrated that CNN features could be utilized for depth inference. Liu~\etal~\cite{liu2016learning} improved the quality of depth by combining a superpixel-based conditional random field (CRF) to a CNN. In contrast to the previous approaches that utilize the ground truth depth as supervision, some works~\cite{garg2016unsupervised,godard2017unsupervised} utilize the task of view synthesis as supervision for single-view depth estimation. This view synthesis approach was extended to unsupervised end-to-end learning of both depth and ego-motion estimation~\cite{zhou2017unsupervised,li2017undeepvo,mahjourian2018unsupervised,yang2018lego}. These methods simultaneously train a depth and a pose network on temporally contiguous frames with a photometric loss between the target and a synthesized nearby view. These approaches are based on rigid motion, so they do not explicitly consider non-rigid motions or they mask out moving object parts.

\subsection{Stereo matching} \quad
Stereo matching estimates depth from a pair of rectified images captured by a stereo rig. Traditional stereo matching typically consists of four steps: matching cost computation, cost aggregation, optimization, and disparity refinement. Recent studies have focused on neural network design that is inspired by the traditional pipeline. The first of these methods was introduced by Zbontar~\etal~\cite{zbontar2016stereo}, which presented a Siamese network structure to compute matching costs based on the similarity of two image patches. Then, they refine by traditional cost aggregation and refinement as post-processing. Later, Mayer~\etal~\cite{mayer2016large} introduced stacked convolution and deconvolution layers for depth estimation. Recent approaches~\cite{liang2017learning,kendall2017end,chang2018pyramid,im2018dpsnet} propose end-to-end networks designed to follow non-learning approaches. Liang~\etal\cite{liang2017learning} utilize both prior and posterior feature consistency to estimate initial depth and refine it. Other works~\cite{kendall2017end,chang2018pyramid} leverage geometric knowledge in building a cost volume and regress disparity using the SoftMax operation in an end-to-end manner. Im~\etal~\cite{im2018dpsnet} generalize the geometric-inspired method to unstructured multiview stereo.

\subsection{Scene flow estimation}

Scene flow is commonly described as a flow field of the 3D motion over the scene points in an image~\cite{vedula1999three}. 
Several works~\cite{vogel2013piecewise,menze2015object,lv2016continuous,behl2017bounding} have studied the problem of estimating not only the rigid scene flow of static objects but also the non-rigid scene flow of dynamic foreground objects. Vogel \etal~\cite{vogel2013piecewise} formulate the 3D scene flow problem as a combination of motion from the scene and planar segments in 3D space. Lv~\etal~\cite{lv2016continuous} present a factor-graph formulation, which is purely continuous, and a pipeline to decompose scene geometry and motion estimation. Menze~\etal~\cite{menze2015object} propose a scene flow estimation method and a large-scale scene flow dataset. Behl \etal~\cite{behl2017bounding} adopt supervised optical flow and instance segmentation, and show the impact of recognition-based geometric cues.

With the rapid development of deep convolutional neural networks (CNNs), the most recent approaches~\cite{yin2018geonet,zou2018df,yang2018every,ranjan2019collaboration} address the joint estimation of rigid and non-rigid geometry. They introduce an unsupervised learning framework that estimates depth, optical flow and ego-motion of a camera in a coupled way. Yin~\etal~\cite{yin2018geonet} propose a two-stage pipeline that estimates the rigid scene flow and object motion. Zou~\etal~\cite{zou2018df} leverage geometric consistency as an additional supervisory signal that measures the discrepancy between the rigid flow and the estimated flow. Yang~\etal~\cite{yang2018every} introduce a holistic 3D motion parser to model the consistency between depth and 2D optical flow estimation. Ranjan~\etal~\cite{ranjan2019collaboration} introduce a new framework that facilitates competition and collaboration between neural networks acting as adversaries.

In contrast to the previous works, we (1) explicitly reason about residual flow to jointly learn camera pose in the dynamic environment, (2) introduce the depth-aware smoothness loss, and (3) design a multi-phase training scheme. In addition, we show that the proposed method can provide reliable speed and direction estimates of moving objects.

\begin{figure*}[t] 
	\centering
    \includegraphics[width=0.99\linewidth]{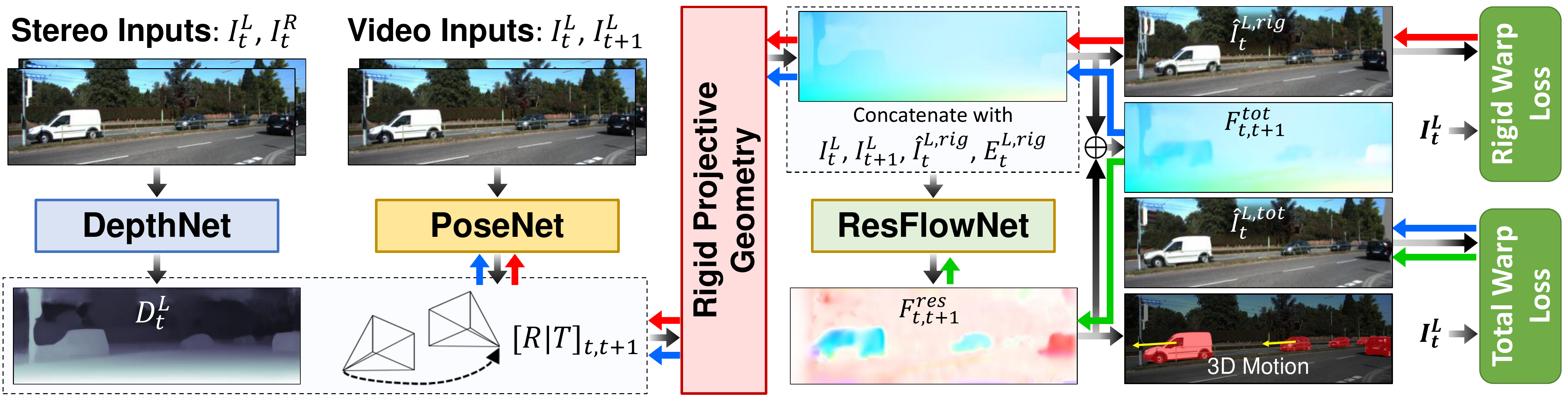}
	\caption{Schematic overview of our method. The proposed method consists of three sub-modules that estimate depth maps, camera poses, and residual flows. We handle the residual flows as dynamic motions of moving objects. The red, green, and blue arrows indicate the directions of gradient back-propagation in optimizing PoseNet and ResFlowNet.}
	\label{fig_overview}
\end{figure*}

%% file: sec_3_v2.tex
\section{PROPOSED METHOD}

Our main objective is to predict the 2D rigid flow of the stationary scene elements and the 3D dynamic motion of moving objects. To achieve this while reducing motion ambiguity, we leverage a stereo vision system with pre-calibrated parameters. We utilize stereo depth to determine the rigid flow between two images based on projective geometry. Then, the proposed residual FlowNet (ResFlowNet) recovers the non-rigid flow, which represents the difference between the rigid flow and total flow and can be attributed to moving objects. In this section, we describe the overall training scheme for the estimation of rigid and residual flow, camera motion, and how to regress the 3D motion of moving objects.

\subsection{System Overview} 
The overall schematic framework of the proposed method is illustrated in~\Figref{fig_overview}.
Our framework has three sub-modules, consisting of depth (DepthNet), camera motion (PoseNet), and residual flow (ResFlowNet) estimation networks. During training, we feed two sequential image pairs ($I^L_{t}$, $I^R_{t}$ which are the left and right images at time ${t}$, and $I^L_{t+1}$, $I^R_{t+1}$ at time $t+1$) into those networks.

\subsection{Depth Estimation} 
For depth estimation from a given pair of stereo images $I^L_{t}$ and $I^R_{t}$, we employ an off-the-shelf algorithm, PSMNet~\cite{chang2018pyramid}, which yields outstanding results on the KITTI stereo benchmark. With a calibrated horizontal focal length $f_x$ and stereo baseline $B$, we convert the stereo disparity $d_t$ into an absolute depth $D_t~(=B\cdot f_x/d_t)$ at metric scale for the following pipeline. We compute the depth map aligned to the left image and denote it as $D^L_t$.

\subsection{Ego-Motion Estimation} 
We first concatenate two sequential left frames $I^L_{t}$ and $I^L_{t+1}$, and pass it through 7 convolution layers to produce an output of 6 channels (3 Euler angles and 3 translation components).
With the predicted depth map and the 6-DoF change in camera pose, the rigid flow $F^{rig}_{t,t+1}$ can be derived from projective geometry as follows:
\begin{equation}
\Scale[0.99]
{
\begin{aligned}
F^{rig}_{t,t+1}(p_t)=K[R|T]_{t,t+1}D_t(p)K^{-1}p_t - p_t ,
\end{aligned}
}
\label{eq_rpg}
\end{equation}
where $p_t$ is the homogeneous coordinates of pixels in $I_t$, and $K$ denotes the camera intrinsic matrix. $R,T$ are a rotation matrix and translation vector, respectively, which are the optimization outputs in this stage.

To train this PoseNet in an unsupervised manner, we impose a photometric loss between a target image $I_{t}$ and a synthesized image $\hat{I}^{rig}_{t}$ warped from $I_{t+1}$ to the target viewpoint by rigid flow $F^{rig}_{t,t+1}$. This synthesis of the rigidly warped image is done using a spatial transformer network~\cite{jaderberg2015spatial}. The photometric error $E^{rig}_t$ for the rigid motion $F^{rig}_{t,t+1}$ is expressed as follows:
\begin{equation}
\Scale[0.99]
{
\begin{aligned}
E^{rig}_t = 
{(1-\alpha)} \left\| { {I_t - {\hat I}^{rig}_t} } \right\|_{1} + {\alpha} \left( 1 - SSIM(I_t, {\hat I}^{rig}_t) \right) ,
\end{aligned}
}
\label{eq_error}
\end{equation}
where $SSIM$ is the structural similarity index~\cite{wang2004image} and $\alpha$ is set to 0.7 based on cross-validation. In this step, we exclude invalid gradients from regions occluded by the rigid transformation through the use of a validity mask~\cite{meister2017unflow}. We set the rigid validity mask ${\bf{M}}^{rig}_t({\rm{x}})$ to be 1 except under the following condition for which it is set to 0:
\begin{equation}
\Scale[0.99]
{
\begin{aligned}
{\left| {F^{rig}_{t,t+1}({\rm{x}}) + F^{rig}_{t+1,t}\left( {{\rm{x}}+F^{rig}_{t,t+1}({\rm{x}})} \right)} \right|^2} \quad\quad\quad\quad\quad\quad\quad\quad \\
> {\gamma _1}\left( {{{\left| {F^{rig}_{t,t+1}({\rm{x}})} \right|}^2} + {{\left| {F^{rig}_{t+1,t}\left( {{\rm{x}} + F^{rig}_{t,t+1}({\rm{x}})} \right)} \right|}^2}} \right) + {\gamma _2} ,
\end{aligned}
}
\label{eq_mask}
\end{equation}
where \rm{x} is the location of each pixel, and we set $\gamma_1=0.01$ and $\gamma_2=0.5$.
Finally, we define the rigid warping loss for PoseNet as follows:
\begin{equation}
\Scale[0.99]
{
\begin{aligned}
\calL_{rig} = 
\sum\limits_{{\rm{x}} \in X} {\mathbf{M}^{rig}_t}({\rm{x}}) \cdot E^{rig}_t({\rm{x}}) .
\end{aligned}
}
\label{eq_loss_rig}
\end{equation}

\begin{figure}[t] 
	\centering
    \includegraphics[width=0.9\linewidth]{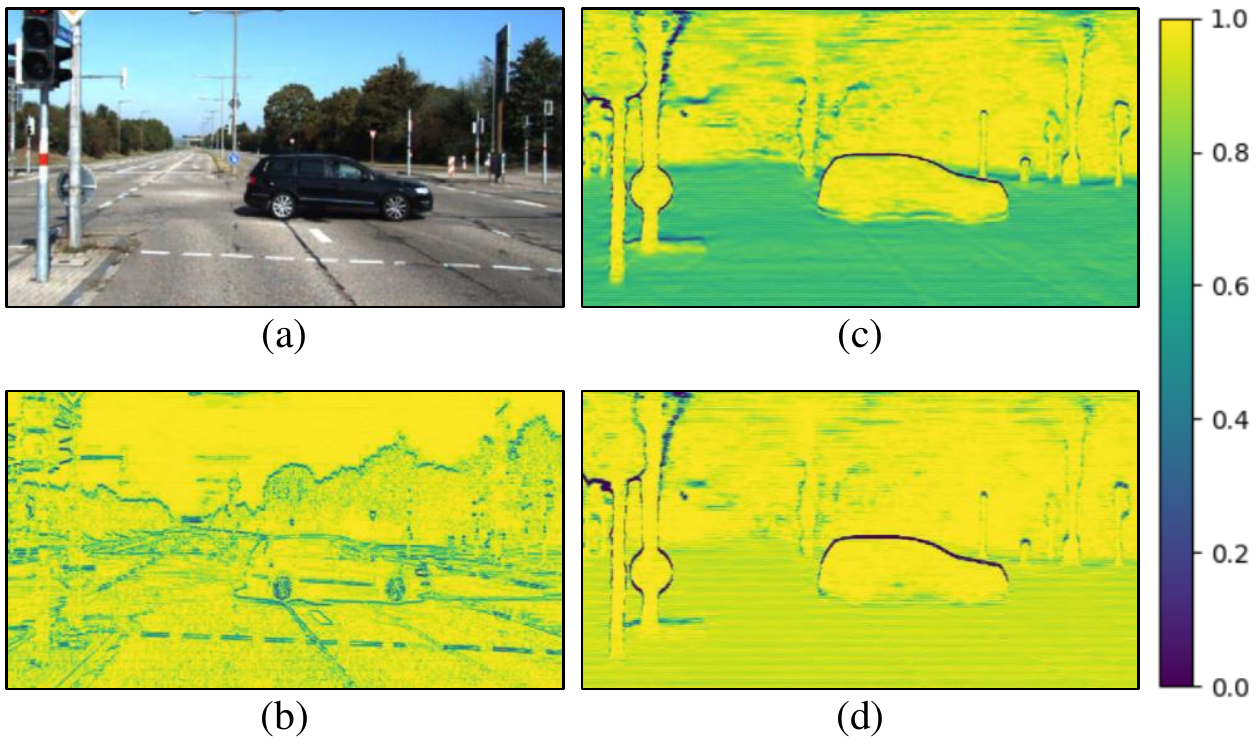}
	\caption{Comparison of different gradient terms in the $y$-direction. (a) Input image. (b) $1^{st}$ order image gradient term, $e^{-| \nabla_{y} I_t(\rm{x}) |}$. (c) $1^{st}$ order depth gradient term, $e^{-| \nabla_{y} D_t(\rm{x}) |}$. (d) $2^{nd}$ order depth gradient term, $e^{-| \nabla^2_{y} D_t(\rm{x}) |}$.}
	\label{fig_grad}
\end{figure}

\subsection{Residual Flow Estimation}
In this subsection, we propose a residual flow network (ResFlowNet) that estimates the 3D flow of moving objects. As input, ResFlowNet takes concatenated data from the sequential images $I_t$, $I_{t+1}$, the warped image $\hat{I}^{rig}_t$, the rigid flow $F^{rig}_{t,t+1}$, and the photometric error $E^{rig}_t$ (12 channels in total). The network adopts the ResNet-50~\cite{he2016deep} structure for encoding.
To infer multi-scale dense flow, we design the decoder to output multi-scale predictions similar to DispNet~\cite{mayer2016large}.
Mid-layer skip-connections are deployed after each residual block as described in~\cite{he2016deep}.
Since ResNet-50 has 4 residual blocks, we predict 4 different scales of output flow maps.

To learn the residual flow in an unsupervised manner, we consider the total flow $F^{tot}_{t,t+1}$ as a sum of the rigid flow $F^{rig}_{t,t+1}$ and a residual flow $F^{res}_{t,t+1}$.
Same as in~\Eqnref{eq_error} and \Eqnref{eq_mask}, we formulate the total synthesis error $E^{tot}_t$ and total validity mask ${\bf{M}}^{tot}_t({\rm{x}})$ by replacing $F^{rig}_{t,t+1}$ with $F^{tot}_{t,t+1}$ in each.
The total warping loss for each scaled prediction is thus defined as follows:
\begin{equation}
\Scale[0.99]
{
\begin{aligned}
\calL_{tot} = 
\sum\limits_{{\rm{x}} \in X} {\mathbf{M}^{tot}_t}({\rm{x}}) \cdot E^{tot}_t({\rm{x}}) .
\end{aligned}
}
\label{eq_loss_tot}
\end{equation}

To mitigate spatial fluctuation, various types of smoothness losses have recently been proposed~\cite{zhou2017unsupervised,mahjourian2018unsupervised,yang2018lego}.
In our stereo setup, the depth network produces an accurate depth map, so we utilize its meaningful boundary information to regularize the 2D flow prediction. Taking advantage of our setup, we propose a depth-aware smoothness loss $\calL_{das}$ as
\begin{equation}
\Scale[0.99]
{
\begin{aligned}
\calL_{das}(D_t, F^{res}_{t,t+1}) = 
\sum\limits_{{\rm{x}} \in X} \sum\limits_{d \in x,y}  \left\| \nabla^2_{d} F^{res}_{t,t+1}(\rm{x}) \right\|_{1} e^{-| \nabla^2_{d} D_t(\rm{x}) |}.
\end{aligned}
}
\label{eq_smooth}
\end{equation}
We input our residual flow and aligned depth map to penalize flow fluctuations except at object boundaries.
This allows similar residual flow values to be assigned to spatially adjacent pixels.
The loss is defined based on the second-order gradients of both depth and residual flow along the $x$- and $y$-directions in 2D space.
We show a visual example of the different characteristics of the $y$-direction gradient terms in~\Figref{fig_grad}.
The image gradient cannot robustly handle low-level color changes, \ie, road markings, which cause noisy boundary information, while the depth gradient can clearly identify object boundaries.
Moreover, we have observed that the second-order depth gradient term has greater effect in penalizing fluctuations on flat and homogeneous regions, \ie, road and wall, than the first-order term. 
Due to limitations of the photometric loss, finding correspondences in homogeneous regions produces a large amount of fluctuation. 
We suppress this problem effectively by using the second-order depth gradient term while training ResFlowNet. 

To sum up, our final loss term through the whole framework is defined as
\begin{equation}
\begin{aligned}
\calL_{n} = \lambda^{(n)}_{rig} \calL_{rig} + \lambda^{(n)}_{tot} \calL_{tot} + \lambda^{(n)}_{das} \calL_{das} ,
\end{aligned}
\end{equation}
where $\Lambda_n=\{\lambda^{(n)}_{rig}, \lambda^{(n)}_{tot}, \lambda^{(n)}_{das}\}$ denotes the set of loss weights in the $n^{th}$ training phase. 


\begin{figure}[t] 
	\centering
    \includegraphics[width=0.99\linewidth]{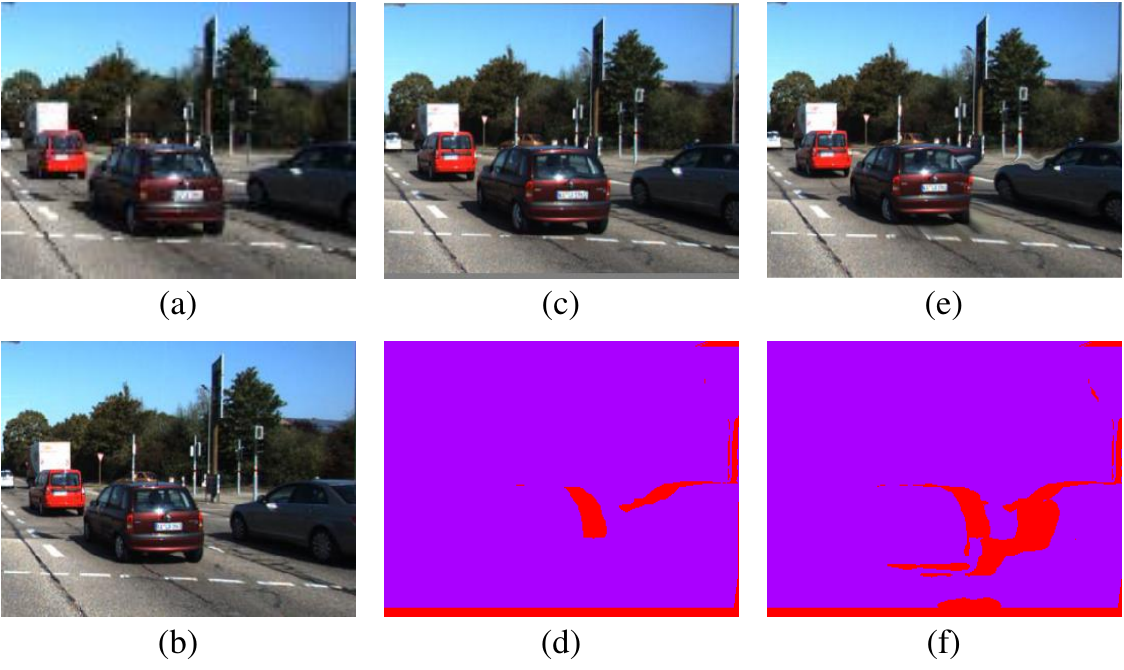}
	\caption{(a) Source frame. (b) Target frame. (c) Warped image by rigid flow. (d) Rigid validity mask. (e) Warped image by total flow. (f) Total validity mask. Red regions on validity masks are outliers when learning ResFlowNet.}
	\label{fig_occ}
\end{figure}

\subsection{Multi-Phase Training}
We propose a multi-phase training scheme to optimize PoseNet and ResFlowNet efficiently considering the characteristics of residual learning. Our system has two back-propagation paths as shown in~\Figref{fig_overview}. The first path is to optimize PoseNet with the directions indicated by red arrows. 
In the first phase of training, we optimize PoseNet solely with the rigid warping loss $\calL_{rig}$. We set the loss weights as $\Lambda_1 = \{0.2, 0, 0\}$ by cross-validation.

In the second training phase, we fix the weights of PoseNet and optimize ResFlowNet solely with the total warping loss $\calL_{tot}$ and the depth-aware smoothness loss $\calL_{das}$. The direction of back-propagation is denoted by green arrows in~\Figref{fig_overview}.
We set the loss weights as $\Lambda_2 = \{0, 1.0, 0.2\}$ by cross-validation.

In the last phase of training, we simultaneously train PoseNet and ResFlowNet. In this step, the new paths of gradient back-propagation are represented by blue arrows (from the total warping loss to PoseNet). 
In the PoseNet optimization, an additional supervisory signal provided by the total warping loss is included. Since flow information decomposed into rigid and residual flow can better reconstruct the target image than only the rigid flow, we conjecture that this new path has a favorable effect on predicting residual flows.
In this last phase, we replace $\mathbf{M}^{rig}_t$ with $\mathbf{M}^{tot}_t$ in~\Eqnref{eq_loss_tot} and set the loss weights as $\Lambda_3 = \{0.1, 1.0, 0.2\}$ by cross-validation.

In our system, the goal is to explicitly reason about the motion of moving objects.
However, in the residual flow field, there are other aspects of pixel displacement, \ie, covering occluded regions.
Those issues are fundamentally caused from the limitations of the photometric loss terms.
To prevent those unnecessary gradient back-propagations as much as possible, we use a two-stage validity mask.
We observe that while training ResFlowNet, the two-stage validity mask also has residual characteristics.
\Figref{fig_occ} shows that the final validity mask filters out additional occluded regions caused by moving objects.

\subsection{2D to 3D Motion Regression}
With the predicted depth, the 2D optical flow vectors can be projected into the 3D space.
From these outputs, instance-level 3D object motion can be predicted.
We utilize the instance information produced by the off-the-shelf algorithm of PANet~\cite{liu2018path}.
We define the instance segmentation mask as $S^{(i)}_t \in \{0,1\}^{H \times W}$ that provides the binary mask information of the $i^{th}$ instance at time $t$.
Then, we can back-project each instance pixel in 2D space into 3D space as follows:
\begin{equation}
\Scale[0.99]
{
\begin{aligned}
\Omega(D_t,p^{(i)}_t)=D_t(p^{(i)}_t)K^{-1}p^{(i)}_t ,
\end{aligned}
}
\label{eq_proj}
\end{equation}
where $p^{(i)}_t$ indicates pixel locations for which $S^{(i)}_t=1$.
According to the generated flow field, the corresponding pixel $\hat p_{t+1}$ at time $t+1$ is $p_t+F^{tot}_{t,t+1}$. 
Therefore, the 3D motion vectors of the $i^{th}$ instance from time $t$ to $t+1$ can be computed as
\begin{equation}
\Scale[0.99]
{
\begin{aligned}
\mathcal{M}^{(i)}_{t,t+1} = [R|T]^{-1}_{t,t+1}\Omega(D_{t+1},p^{(i)}_t+F^{tot}_{t,t+1}) - \Omega(D_t,p^{(i)}_t) .
\end{aligned}
}
\label{eq_motion}
\end{equation}
We employ the RANSAC scheme to estimate a representative magnitude and direction among the 3D motion vectors of the $i^{th}$ instance.

%% file: sec_4.tex
\section{EXPERIMENTS}
We evaluate the performance of our system and compare with previous supervised and unsupervised methods on optical flow and visual odometry.
We train and test our method on the KITTI dataset~\cite{geiger2012we} for benchmarking.

\subsection{Implementation Details}
\textbf{Training}~~Our system is implemented in PyTorch~\cite{paszke2017automatic}. We train our networks using the ADAM optimizer~\cite{kingma2015adam} with $\beta_1 = 0.9$ and $\beta_2 = 0.999$ on $8\times$Nvidia Titan X GPUs. The initial learning rate is set to $2 \times 10^{-4}$ for the first and second training phases, and $10^{-4}$ for the third training phase. During training, we use a batch size of 4 and decrease each learning rate by half every 30 epochs. We train PoseNet for about 70 epochs in the first phase and train ResFlowNet for about 40 epochs in the second phase. In the third phase, we train both networks until the validation accuracy (evaluated with flow ground truth) converges. 

\textbf{Networks}~~For the stereo matching network, we use the $Stacked Hourglass$ model of PSMNet pretrained with the KITTI 2015 stereo dataset. Since the encoding part of ResFlowNet is adopted from ResNet-50, we use the ImageNet~\cite{russakovsky2015imagenet} pretrained model as initialization of ResFlowNet except for the first convolutional layer, which receives an input of 12 channels. 
The number of parameters and running time of each sub-module are listed in Table~\ref{tab_spec}.

\begin{table}[t]
\huge
\renewcommand{\tabcolsep}{5mm}
\centering
\caption{Number of parameters and forward pass time of each network.}
\vspace{0mm}
\label{tab_spec}
\begin{adjustbox}{width=0.40\textwidth}
\begin{tabular}{lcc}
\toprule
Sub-module & No. params. & Runtime ($ms$) \\
\midrule
Stereo Matching~\cite{chang2018pyramid} & 5,224,768 & 410 \\
ResFlowNet                              & 73,403,272 & 35 \\
PoseNet                                 & 1,586,774 & 5 \\
\midrule
Total           & 80,214,814 & 450 \\
\bottomrule
\end{tabular}
\end{adjustbox}
\end{table}

\textbf{Dataset}~~
The proposed network is learnable in an unsupervised manner, and we train our models with KITTI raw stereo videos. 
In the joint training of PoseNet and ResFlowNet, we take two consecutive stereo pairs as input: stereo pairs for stereo matching inference, and two sequential left images.
This results in a total of 20,000 paired training sequences.
We pre-process the input image size to be $832 \times 256$ for each sub-module, which is twice as large as the input in \cite{zhou2017unsupervised}, for more detailed inference.
We perform data augmentation by left-right flipping and reversing the time direction.

\subsection{Optical Flow Estimation}
To evaluate performance on optical flow estimation, we test our networks with the KITTI 2015 scene flow training set, which has 200 urban scenes captured by stereo RGB cameras and a Velodyne laser scanner, with sparse 3D scene flow ground truth generated by point cloud projection.
For a fair evaluation, we exclude scenes from the training set that also appear in the test set.

In this evaluation, we compare our method with state-of-the-art supervised optical flow methods (\eg, FlowNet~\cite{dosovitskiy2015flownet}, FlowNet2~\cite{ilg2017flownet}, SpyNet~\cite{ranjan2017optical}, and PWC-Net~\cite{sun2018pwc}), and unsupervised methods (\eg, DSTFlow~\cite{ren2017unsupervised}, UnFlow~\cite{meister2017unflow}, OccAwareFlow~\cite{wang2018occlusion}, GeoNet~\cite{yin2018geonet}, DF-Net~\cite{zou2018df}, and EPC~\cite{yang2018every}). 
For the supervised methods, we report their performance without finetuning.
Our evaluation employs common quantitative measures for optical flow and scene flow, namely endpoint error (EPE) and F1-score.

\begin{figure*}[t] 
    \centering
    \includegraphics[width=0.99\linewidth]{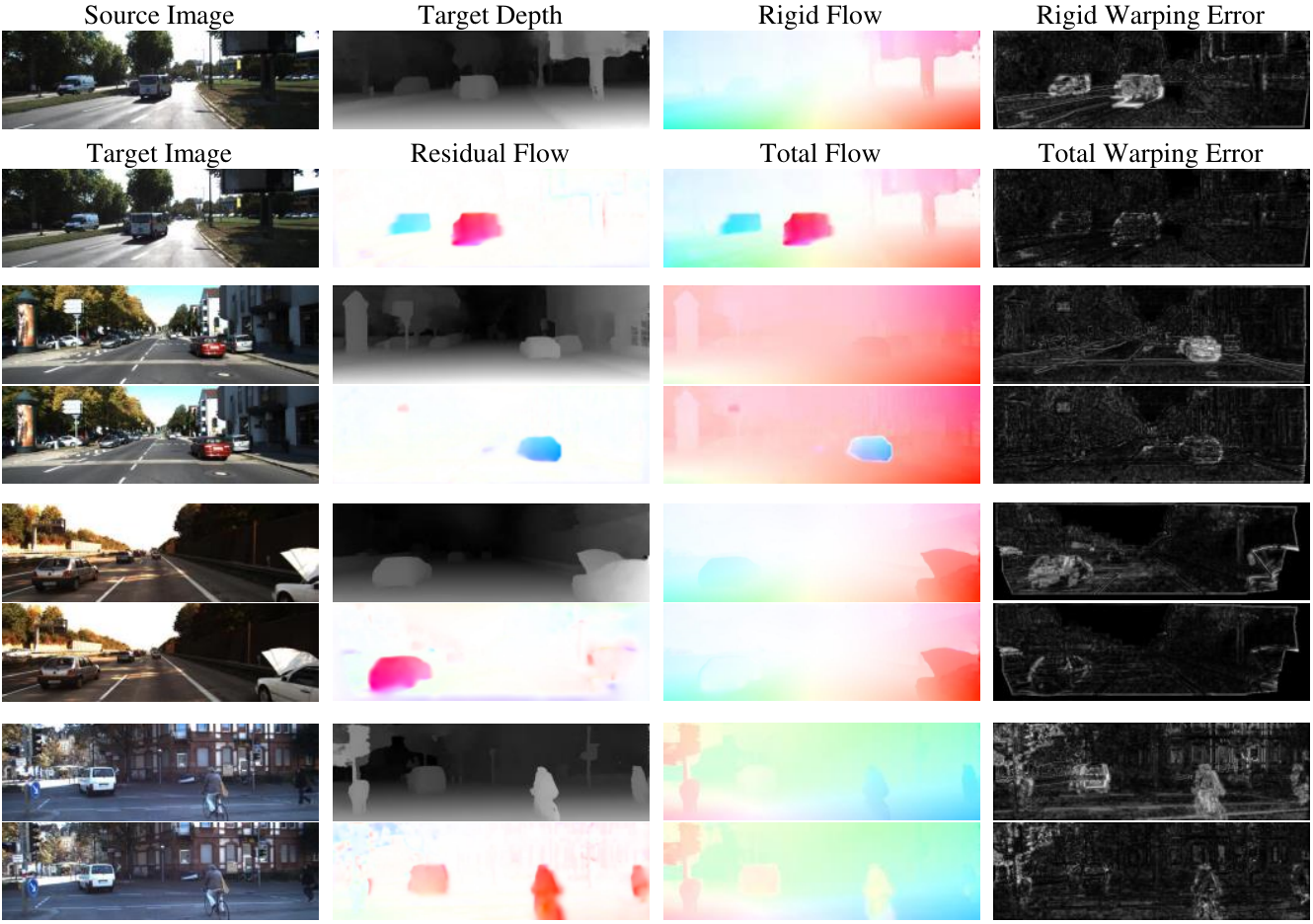}
    \vspace{-2mm}
    \caption{Qualitative results of optical flow estimation, which show that the rigid warping errors after the $2^{nd}$ training phase are significantly refined by the $3^{rd}$ training phase.}
    \label{fig_flow}
\end{figure*}

As shown in~\Tabref{tab_flow}, the proposed network achieves the lowest EPE and F1-score in overall pixels and the second best in non-occluded regions. 
An ablation study of the proposed multi-phase training scheme shows that the joint learning of PoseNet and ResFlowNet (ours, $3^{rd}$ phase) improves the performance significantly compared to separate learning (ours, $2^{nd}$ phase). 
This shows that gradient back-propagation while learning camera pose produces a favorable effect on learning residual flow.
The comparison between ours and GeoNet, which has a similar architecture of ResNet-50 but no joint training, also supports this reasoning.
Compared to UnFlow-C, the average EPE of ours is slightly higher, but the F1-score shows noticeable improvements from our method.
According to this result, we conjecture that ours is robust on occluded regions which are additionally generated by residual flows from dynamic motions since the ratio of our correctly estimated flows (including occlusion) is higher than others.
Among the methods using geometric constraints, ours demonstrates the best result on flow estimation. 
This shows that using the residual characteristics of rigid (geometric) and non-rigid (non-geometric) motion has considerable benefits for 2D matching.
The qualitative results in~\Figref{fig_flow} shows that our method effectively decouples the rigid and non-rigid flows.

\begin{table}[t]
\renewcommand{\tabcolsep}{1mm}
\centering
\caption{Optical flow evaluation on the KITTI scene flow 2015 training set. F1: F1-score representing the percentage of erroneous pixels. EPE: average endpoint error, where a pixel is considered to be correctly estimated if the flow EPE is $<3\text{px}$ or $<5\%$. For both metrics, lower is the better. The datasets used are FlyingChairs (C)~\cite{dosovitskiy2015flownet}, Sintel (ST)~\cite{butler2012naturalistic}, FlyingThings (T)~\cite{mayer2016large}, KITTI (K), and SYNTHIA (SY)~\cite{ros2016synthia}. Noc denotes non-occluded pixels.}
\vspace{0mm}
\label{tab_flow}
\begin{adjustbox}{width=0.47\textwidth}
\begin{tabular}{@{}lccccccc@{}}
\toprule
\multirow{2}{*}{Method} & & & & & Noc & \multicolumn{2}{c}{All} \\
\cmidrule(l{2pt}r{2pt}){6-6} \cmidrule(l{2pt}r{2pt}){7-8}
 & dataset & stereo & geometry & supervised & EPE & EPE & F1 (\%) \\
\midrule
FlowNetS~\cite{dosovitskiy2015flownet}  & C+ST &  &  & \checkmark & 8.12 & 14.19 & -- \\
FlowNet2~\cite{ilg2017flownet}          & C+T  &  &  & \checkmark & 4.93 & 10.06 & 30.37 \\
SPyNet~\cite{ranjan2017optical}         & C+T  &  &  & \checkmark & -- & 20.26 & -- \\
PWC-Net~\cite{sun2018pwc}               & C+T  &  &  & \checkmark & -- & 10.35 & 33.67 \\
\midrule 
DSTFlow~\cite{ren2017unsupervised}    & K     &            &            &  & -- & 16.79 & 36.00 \\
UnFlow-C~\cite{meister2017unflow}     & K     &            &            &  & \textbf{4.29} & 8.80 & 28.95 \\ 
OccAwareFlow~\cite{wang2018occlusion} & K     &            &            &  & -- & 8.88 & -- \\ 
GeoNet~\cite{yin2018geonet}           & K     &            & \checkmark &  & 8.05 & 10.81 & --\\
DF-Net~\cite{zou2018df}               & K+SY  &            & \checkmark &  & -- & 8.98 & --\\
EPC~\cite{yang2018every}              & K     & \checkmark & \checkmark &  & -- & -- & 25.74\\
\midrule 
Ours ($2^{nd}$ phase) & K & \checkmark & \checkmark &  & 6.91 & 9.98 & 24.77 \\
Ours ($3^{rd}$ phase) & K & \checkmark & \checkmark &  & 4.68 & \textbf{8.74} & \textbf{20.88} \\
\bottomrule
\end{tabular}
\end{adjustbox}
\end{table}

\begin{table}[t]
\huge
\renewcommand{\tabcolsep}{5mm}
\centering
\caption{Absolute Trajectory Error (ATE) on the KITTI odometry dataset.}
\vspace{0mm}
\label{tab_odom}
\begin{adjustbox}{width=0.47\textwidth}
\begin{tabular}{lccc}
\toprule
Method & No. frames & Seq. 09 & Seq. 10 \\
\midrule
ORB-SLAM (full)~\cite{mur2015orb}           & All & $0.014\pm0.008$ & $0.012\pm0.011$ \\
ORB-SLAM (short)~\cite{mur2015orb}          & 5   & $0.064\pm0.141$ & $0.064\pm0.130$ \\
SfM-Learner~\cite{zhou2017unsupervised}     & 5   & $0.021\pm0.017$ & $0.020\pm0.015$ \\ 
SfM-Learner (updated)~\cite{zhou2017unsupervised}    & 5   & $0.016\pm0.017$ & $0.013\pm0.015$ \\ 
Vid2Depth~\cite{mahjourian2018unsupervised} & 3   & $0.013\pm0.017$ & $0.012\pm0.015$ \\
GeoNet~\cite{yin2018geonet}                 & 3   & $0.012\pm0.007$ & $0.012\pm0.009$ \\ 
\midrule
Ours ($1^{st}$ phase) & 2   & $0.013\pm0.007$ & $0.012\pm0.023$ \\ 
Ours ($3^{rd}$ phase) & 2   & $\mathbf{0.012\pm0.006}$ & $\mathbf{0.012\pm0.007}$ \\
\bottomrule
\end{tabular}
\end{adjustbox}
\vspace{0mm}
\end{table}

\subsection{Pose Estimation}

Performance on pose estimation is compared to state-of-the-art methods (\eg, ORB-SLAM~\cite{mur2015orb}, SfM-Learner~\cite{zhou2017unsupervised}, Vid2Depth~\cite{mahjourian2018unsupervised}, and GeoNet~\cite{yin2018geonet}) on the KITTI visual odometry dataset. 
For a fair comparison, we directly report the results of comparison methods from~\cite{yin2018geonet,mahjourian2018unsupervised} in~\Tabref{tab_odom}. 
Following their evaluation setting, we measure the absolute trajectory error (ATE).

An ablation study of our method compared to only the $1^{st}$ training phase shows that the joint learning of PoseNet and ResFlowNet (ours, $3^{rd}$ phase) has a positive effect on learning ego-motion. 
We conjecture that the better reconstruction of a target image from the total warping loss reduces the outlier errors which rigid projective geometry cannot solve. 
This leads to better gradient information for learning PoseNet.
Continuing the ablation study in~\Tabref{tab_flow}, we show the effectiveness of the proposed multi-phase training scheme on learning residual flow.

In contrast to~\cite{yin2018geonet}, we have not finetuned our model using the KITTI visual odometry dataset. Moreover, we use only 2 frames to regress the camera poses while all competing methods use more than 3 images. As shown in~\Tabref{tab_odom}, our method outperforms all of the competing baselines. 

\begin{figure}[t] 
    \centering
    \includegraphics[width=0.99\linewidth]{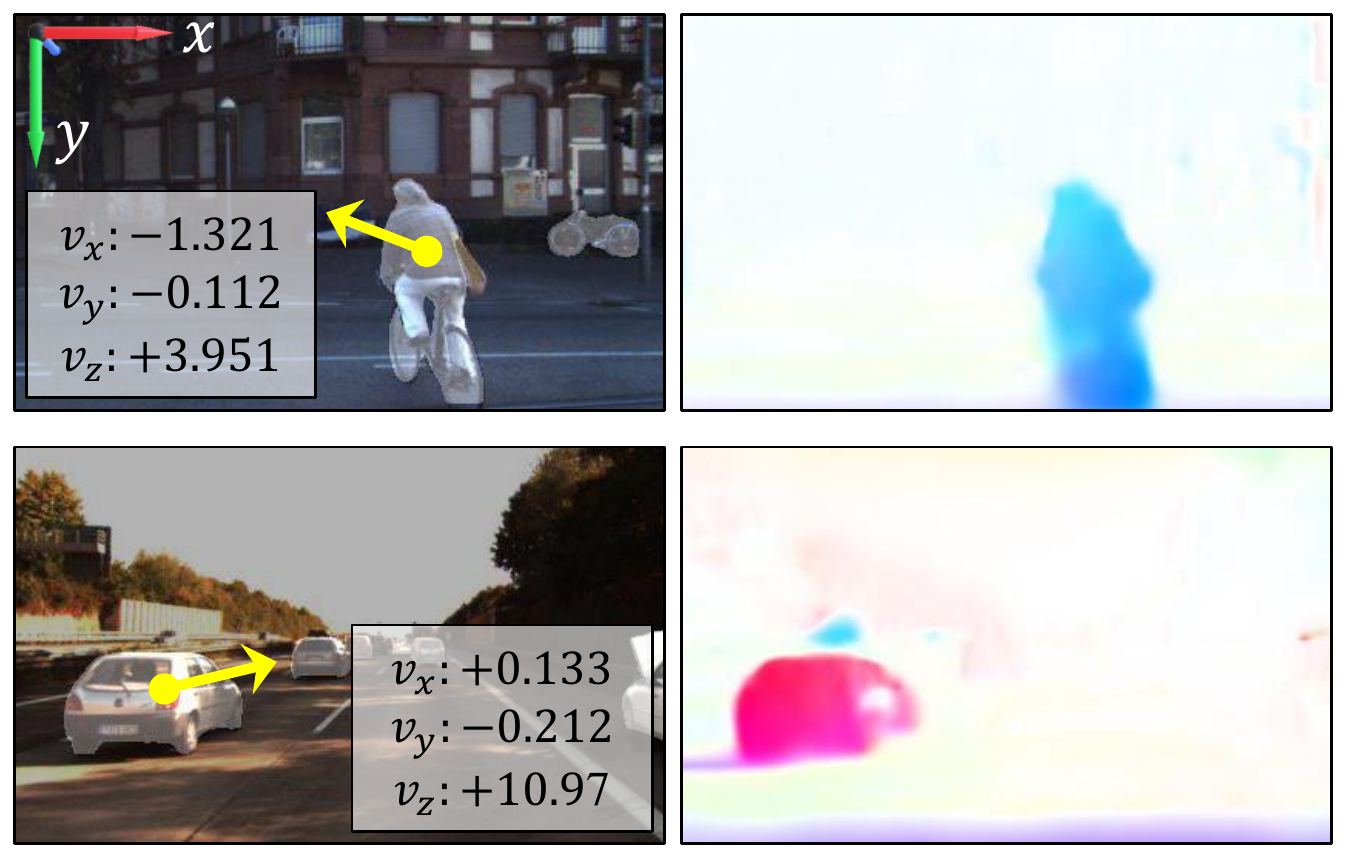}
    \vspace{-4mm}
    \caption{Images on the left are 3D motion predictions of moving objects while the camera moves. Images on the right are the corresponding residual flows. Dividing motion vectors by the camera sampling time (KITTI: 0.1~$sec$) gives the velocity~($m/s$) of dynamic objects.}
    \label{fig_motion}
\end{figure}

\subsection{3D Motion Prediction}

Qualitative results of our method on 3D motion prediction of moving objects are illustrated in~\Figref{fig_motion}. Our system predicts the speed and movement direction of all instances together with the rigid flow and camera motion. The 3D motion vectors are represented using the right-handed coordinate system ($x$-axis: right, $y$-axis: down, $z$-axis: forward), same as the KITTI machine vision camera coordinates~\cite{menze2015object}. We believe that the outputs of the proposed method can be beneficially used in a number of robotics applications.